# HFN: Heterogeneous Feature Network for Multivariate Time Series Anomaly Detection

Jun Zhan

*College of Computer Science, National University of Defense Technology, Changsha 410073, China*

Chengkun Wu[*]

*State Key Laboratory of High Performance Computing, College of Computer Science, National University of Defense Technology, Changsha 410073, China*

Canqun Yang

*National SuperComputing Center in Tianjin, China*

Qiucheng Miao

*College of Computer Science, National University of Defense Technology, Changsha 410073, China*

Xiandong Ma*

*School of Engineering, Lancaster University, Lancaster, LA1 4YW, UK*

**Abstract:** Network or physical attacks on industrial equipment or computer systems may cause massive losses. Therefore, a quick and accurate anomaly detection (AD) based on monitoring data, especially the multivariate time-series (MTS) data, is of great significance. As the key step of anomaly detection for MTS data, learning the relations among different variables has been explored by many approaches. However, most of the existing approaches do not consider the heterogeneity between variables, that is, different types of variables (continuous numerical variables, discrete categorical variables or hybrid variables) may have different and distinctive edge distributions. In this paper, we propose a novel semi-supervised anomaly detection framework based on a heterogeneous feature network (HFN) for MTS, learning heterogeneous structure information from a mass of unlabeled time-series data to improve the accuracy of anomaly detection, and using attention coefficient to provide an explanation for the detected anomalies. Specifically, we first combine the embedding similarity subgraph generated by sensor embedding and feature value similarity subgraph generated by sensor values to construct a time-series heterogeneous graph, which fully utilizes the rich heterogeneous mutual information among variables. Then, a prediction model containing nodes and channel attentions is jointly optimized to obtain better time-series representations. This approach fuses the state-of-the-art technologies of heterogeneous graph structure learning (HGSL) and representation learning. The experiments on four sensor datasets

---

[*] Corresponding author
Email address: *chengkun_wu@nudt.edu.cn* (Chengkun Wu), *xiandong.ma@lancaster.ac.uk* (Xiandong Ma)

from real-world applications demonstrate that our approach detects the anomalies more accurately than those baseline approaches, thus providing a basis for the rapid positioning of anomalies.

## 1. Introduction

As industrial internet of things technology develops, an increasing number of industrial systems is exposed to the internet, posing serious risks to their ability to operate securely [1]. Continuous monitoring the operation data of the system and precisely and effectively identifying potential attacks or the evolution of the equipment condition by use of these data is an effective technique to handle these challenges [2, 3]. For instance, an operation and maintenance personnel in a large power plant can quickly identify abnormal sensor behavior using the precise intrusion detection systems, which are developed by massive amounts of data collected by the supervisory control and data acquisition (SCADA) system [4, 5], providing them a possibility to prevent potential system failures before irreversible damage. However, these monitoring data always have complicated structures, high dimensionality, and hard labeling, making manual tasks difficult to handle. Therefore, it is vitally necessary to investigate the semi-supervised or unsupervised time-series anomaly detection approach by utilizing a sizable amount of complicated unlabeled data.

Recently, deep learning technique has been applied successfully in various anomaly detection problems [6]. For high-dimensional MTS analysis, the temporal relations between different timestamps are considered first [7]. Because of their capability of capturing long-term dependency relations, recurrent neural network [8] and temporal convolutional network [9] were demonstrated to achieve better results on the time-series tasks involving single or multiple variables [10]. However, various sensors could be mutually coupled. The capacity of these approaches to detect abnormalities may be constrained by their modeling of solely temporal variables. Therefore, it is crucial to take into account both the temporal features of different timestamps and potential correlations among these variables [11, 12]. Combining the sequential network and the convolution neural network (CNN) is an effective way to achieve this. Cross-correlation among high-dimensional data can be extracted by using the local perception capacity of the convolutional kernel [13]. However, CNN is primarily used to handle Euclid-space data, such as image [14]. There exist some limitations on the MTS with different attributes. In such cases, the graph neural network (GNN) has been successfully applied into the modelling of MTS due to its good structure modelling capability between complex data; the most advanced results are achieved in [11, 15].

With regards to the latent feature modeling of time-series data, the variable attributes from the data are generally seen as homogeneous in the most existing papers; that is, the data types are treated without distinction, such as use of the variational autoencoders [16] and generative adversarial networks [17]. These methods model complex distribution from large-scale high-dimensional datasets. After the training is finished by using the dataset from normal conditions, the similar generative data are viewed as normality, while the dissimilar data are viewed as anomalies. However, there are still fewer works considering the heterogeneity of time-series data, although this kind of data are abundant in practical situations. For instance, in a large-scale water processing system [17], the information, such as flow, pressure and liquid level collected by the sensors in the intermediate process, is collected as the numeric continuous values. However, the signals, such as valve state and location collected by the sensors of the actuator, are generally the categorical discrete values.

Inputting the mixed type of heterogeneous data into a deep learning network may cause the useful information to be ignored and therefore satisfied results cannot be obtained. The fundamental reason is that there are totally different edge distributions between the variables with different types [18, 19].

To overcome the limitation of deep learning model in such circumstances, we propose a heterogeneous feature learning network for MTS, and study its abnormal detection capability with the extensive real-world datasets. The overall framework can be divided into three stages: 1) Heterogeneous graph structure learning (HGSL) stage for MTS. We fuse the sensor embedding vector similarity matrix and the feature value similarity matrix of different variable categories to model the heterogeneous structural information. Moreover, we propose a category-based fixed-length approach to replace the widely used meta-path [20] for extracting heterogeneous relation subgraphs. 2) Heterogeneous representation learning stage for MTS. We embed different kinds of variables into vectors for fusion. Distinct from the previous heterogeneous graph attention network [21], we further expand the channel attention on the basis of node attention and semantic attention, so as to achieve a joint optimization training of node embedding representation with different types. 3) Abnormal detection and location stage. By analyzing the deviation between the predicted and real values, we calculate a condition score for each sensor, where the largest condition score is considered as the maximum abnormal probability.

The major contributions of the paper are summarized as follows:
1) We propose a novel HGSL approach for MTS, which learns heterogeneous graph structure information between sensor-embedding vectors and category-based feature value vectors simultaneously.
2) We propose a heterogeneous feature network (HFN) and apply it to MTS anomaly detection. Our approach successfully learned the dynamic dependency among different variables and timestamps by utilizing two single-level attention mechanisms, namely attention-based node embedding and channel aggregation,
3) The extensive experiments indicate that HFN can detect the anomalies from real-world MTS datasets and is proved to outperform the most existing methods. Besides, we analyze the condition scores of MTS, demonstrating that the proposed method has the advantage of locating the anomalies.

The rest of this paper is structured as follows. Section 2 describes the related work of MTS anomaly detection. Section 3 presents the structure and working principle of HFN-based MTS anomaly detection framework in detail. Section 4 show the performance of proposed method on three real-world MTS datasets. Finally, the conclusion and future improvements are given in Section 5.

## 2. Related work

MTS anomaly detection has extensive application prospects in the fields of industry, financial business, and the Internet of Things. As the key research problem in this paper, we firstly review the related work for MTS anomaly detection, which can generally be categorized as unsupervised, supervised, and semi-supervised. We focus on studying data heterogeneity modeling of MTS, especially heterogeneity representation learning from time-series data, graph structure learning, and heterogeneous graph neural network.

## 2.1. MTS anomaly detection

MTS anomaly detection is typically regarded as an unsupervised learning problem [22], and algorithms based on clustering [23], such as K-means [24], fuzzy c-means [25], or spatiotemporal clustering [26], are frequently used. By grouping time-series data into various clusters, these techniques can identify anomalies by calculating the similarity or distance between the observed value [27] and the cluster center [28]. However, unsupervised detection methods usually focus more on static data model development. In contrast, a supervised abnormal detection algorithm has a higher detection accuracy. Under the circumstance of high-quality labeling, the indicator accuracy can be approximate to 100% [29]. However, the supervised detection requires that the training set contains correctly both labeled positive and negative samples, which is often not easy. [30]. Fortunately, in the actual cases, we have a chance to obtain a large quantity of data under the normal conditions [17], making the semi-supervised abnormal detection attract wide attentions [31]. In the latest work, Miryam et al. [32] proposes the methods to show the great advantages and extensive application prospect of the semi-supervised algorithm in MTS abnormal detection.

## 2.2. Modeling for heterogeneous data

The data heterogeneity has been widely concerned such as in the music recommendation system [33], academic network [34] and social platform [35]. Heterogeneous learning method usually focuses on capturing and integrating couplings with multiple variable types in the same or different levels. To learn the embedding representation of heterogeneous data, the matrix decomposition method is traditionally adopted [36, 37]. However, it is usually very expensive and low-efficient in terms of the computation cost of decomposing a large-scale matrix [38]. Moreover, the discretization of continuous features [39] or continuous data [40] are also a typical method; however this transformation may ignore the correlation between variables. To solve these challenges, heterogeneous graph embedding or heterogeneous graph representation learning [41] has been widely studied. Its main goal is to map the input data into low-dimensional space while simultaneously preserving the heterogeneous structure and semantic characteristics of the data [42]. For instance, for the tasks of text classification, Wang et al. [21] proposed a heterogeneous graph attention network (HAN), which aggregates the features of meta-path based neighbors through a hierarchical manner to generate the embedding representation of nodes. Fu et al. [43] proposed a meta-path aggregated graph neural network (MAGNN) by designing multiple candidate encoder functions to extract heterogeneous information from the meta-path. Wang et al. [44] combined the heterogeneous graph neural network with comparison learning, and proposed a self-supervised heterogeneous graph neural network from both heterogeneous network and meta-path for learning node embedding representation. In the social or citation network, in order to capture the dynamic performances of heterogeneous graph, Hu et al. [45] proposed a heterogeneous graph transformer (HGT) by introducing a relative temporal encoding technique for solving the problem where the dynamic result dependence is difficult to capture. Yang et al. [46] proposed a dynamic heterogeneous graph (DyHAN) utilizing structural heterogeneity and time revolution to learn node embedding.

## 2.3. Graph structure learning

MTS usually exist in the form of tabular data [47], lacking of predefined graph structure required for graph neural network [15], which constitutes the challenge for the modelling [48].

Hence, it is extremely vital to learn the links between edges and refine the graph from the existing time-series data [49]. The existing methods can mainly be divided into three categories: metric-based approaches usually implemented by using kernel function [50, 51], cosine similarity [42, 52, 53] or inner product [54] to calculate the similarity between nodes as edge weights. Neural networks-based approaches have generally utilized a complex deep neural network to model the edge weights of the given node features and representations. For instance, Luo et al. [55] proposed a multilayer perception-based graph structure optimization approach, where the edge number of sparse graph is punished through parameterized network for pruning the edges which are unrelated with the tasks. Zhao et al. [11] proposed a graph structure learning approach with attention coefficient, while Sun et al. [56] utilized a dot-product self-attention to model the dynamic connection relations between the nodes. Direct learning approaches, regarding adjacent matrix as a learnable parameter, make associative learning together with the follow-up tasks for optimization. For instance, Gao et al. [57] proposed the graph learning neural networks (GLNNs) utilizing spectral graph theory for graph learning. However, these approaches mostly aim at learning isomorphic graph structure. To enable capture the heterogeneity between the data efficiently, Zhao et al. [42] proposed a heterogeneous graph learning approach utilizing the fusion of feature similarity sub-graph, feature propagation graph and semantic graph, which successfully learns an appropriate graph structure for a heterogeneous graph neural network.

## 3. Proposed Framework

### 3.1. Problem statement

**Definition 1. MTS Anomaly Detection**

Generally, we define heterogeneous MTS dataset as a time-series dataset with $L$ variables, $N$ different types of sensors, and $T$ length, which is expressed as $X = \{x_{1:T}^N\}$, where $N \in \{type^1, \cdots, type^n\}$ denotes the set of data types. Note that the variable number contained in the specified categories may be larger than 1. For instance, for arbitrary data type $type^n$, all time series at the moment $t$ can be denoted as $x_t^{type^n} \in \{x_t^{type_i^n}, for\ i \in \{0, \cdots, d\}\}$, where $d$ represents the number of time-series sequence in this category. In this paper, we adopt the sliding window-based model training approach. At the moment $t$, we sample a continuous subsequence with the length of $\omega$ as the model input, denoted as $S^N(t) = [x_{t-w+1}^N, \cdots x_t^N]$. For the abnormal detection task, our target is to predict the value of all sensors $x_{t+1}^N$ at the moment $t+1$ by utilizing the input subsequence $S^N(t)$, and obtain the predicted value $\hat{x}_{t+1}^N$. The mean square error (MSE) between the predicted value and practical value is used as loss to optimize the model. According to the usual semi-supervised abnormal detection methods, in the training stage, only the data collected from normal conditions are chosen. However, in the testing stage, the deviation between the predicted value and practical value is further used for calculating the condition scores of the data, while the scores of the corresponding data over the threshold is judged as the anomalies, otherwise normal.

**Definition 2. Graph-based Heterogeneous Feature Learning for MTS**

Specially, we divide time-series data into three data types, that is $N \in \{C, CD, D\}$:
1) Continuous numerical variables $C$, where the value of data are taken from continuous real number, such as $x_t^{C_i} \in \mathbb{R}$.
2) Discrete categorical variables $D$, where the value of data are taken from a limited set of

values, such as $x_t^{D_i} \in \{0, 1, 2\}$.
3) Hybrid variables $S_t^{CD}$ which contain both numerical and categorical variables where the values of the element are taken from the above two categories.

We construct a heterogeneous dynamic graph to model the above MTS. Different time-series variables are viewed as the node in the graph, while their connection relation is seen as the edge. This dynamic graph can be denoted as $\mathbb{G}_{S^N(t)} = (\mathcal{V}, \mathcal{E})$, where $\mathcal{V}$ and $\mathcal{E}$ represent node and edge set respectively. We respectively extract categorical feature subgraph $\mathbb{G}_{S^D(t)}$, numerical feature subgraph $\mathbb{G}_{S^C(t)}$, and categorical and numerical mixed subgraph $\mathbb{G}_{S^{CD}(t)}$ for learning heterogeneous information. For the arbitrary subgraph, its adjacent matrix is $A_N \in \mathbb{R}^{|\mathcal{V}_N| \times |\mathcal{V}_N|}$, where $\mathcal{V}_N$ represents the node set with the specific type. If there exist connection relations between two arbitrary nodes in the subgraph, the corresponding element of adjacent matrix is 1. Noted that the final node embedding integrates the node embedding representations of three different subgraphs.

## 3.2. Model Architecture

Our HFN-based approach aims at learning the complex correlation between different types of time-series data carried by the defined dynamic graph above. For each node, the potential temporal correlation is allowed to be considered with a sliding window along the dataset.

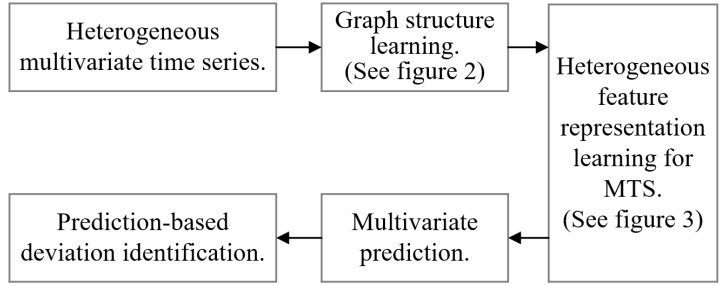

**Figure 1** Architecture of HFN-based MTS anomaly detection framework

**Figure 1** shows the proposed HFN-based semi-supervised abnormal detection framework architecture. It can be seen that for a given MTS, we firstly learn a heterogeneous dynamic graph representing the structural information between different variables (as shown in **Figure 2**), decomposing the time-series data into different graph structures. On this basis, the categorical feature subgraph, the continuous numerical feature subgraph and the hybrid subgraph are extracted and then inputted into the HFN network based on graph attention function to learn the potential embedding representations of each sensor (as shown in **Figure 3**). Then we predict the future values of each sensor based on these embedding representations. Finally, the deviation between the predicted and practical values is used for measuring and locating the anomalies.

## 3.3. Graph structure learning pipeline

To learn the complex heterogeneous potential features between different types of sensors, a key process is how to map the variable correlation from MTS into the adjacent matrix of the graph. In the previous studies, all assumed that the constructed graph is the static isomorphic graph, thus resulting in the loss of some key information. For instance, the significance of variables exists great difference at the operating condition of full-load and partial-load of generating equipment [58]. Hence, as shown in **Figure 2**, we learn the potential heterogeneous graph structure of MTS from the perspectives of global semantic correlation and local feature correlation. For the global semantic correlation, we introduce a learnable embedding vector for each variable, and denote it as $e_i \in \mathbb{R}^{1 \times \omega'}$. For $i \in \{0, \cdots, L\}$, where $\omega'$ represents the dimension of embedding vector. This vector can be learned together with subsequent prediction network parameters. For the local feature correlation, we calculate the potential structural information based on the feature values of the variables. We adopt a special mapping network to project different types of input feature vector $S^N(t)$ into a public space. Taking data type $C$ as an example, the projected feature of the arbitrary variable $x^{C_i}$ is denoted as $f^{C_i} \in \mathbb{R}^{1 \times \omega'}$:

$$f^{C_i} = SELU(x^{C_i} \cdot \boldsymbol{W^C} + \boldsymbol{b^C}) \tag{1}$$

where $x^{C_i} \in \mathbb{R}^{1 \times \omega}$ is the subset of all continuous numerical variables. $\omega$ is the time length of input feature vector. $\boldsymbol{W^C} \in \mathbb{R}^{\omega \times \omega'}$ is learnable weight matrix, and $\boldsymbol{b^C} \in \mathbb{R}^{1 \times \omega'}$ is biasing. Similarly, we can calculate and obtain the projected feature representation $f^{D_i}$ of the discrete categorical variables.

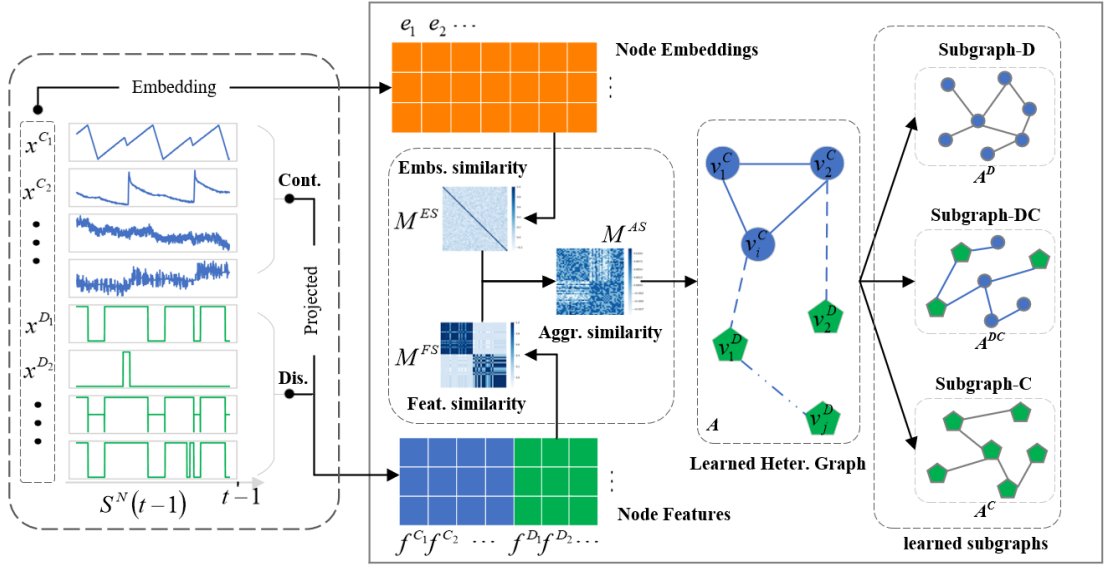

**Figure 2 Structure learning of MTS heterogeneous dynamic graph**

### 3.4. Similarity Graphs

The main task of graph structure learning is to learn an adjacent matrix representing the mutual connection between nodes in the graph. Therefore, we propose a learning approach based on aggregating cosine similarity. According to the embedding for the variables and the mapping of variable feature vectors, we obtain the global semantic embedding matrix $\boldsymbol{E} \in \{e_1, \cdots e_L\}$ and local feature vector representation matrix $\boldsymbol{F^N} \in \{f^{N_1}, \cdots, f^{N_L}\}$. Clearly, these obtained matrixes from different perspectives contain different information. Specifically, we firstly calculate cosine similarity between the elements in different matrixes to obtain their connection information. After obtaining the node embedding (NE) similarity matrix $M^{E_s} \in \mathbb{R}^{L \times L}$ and node feature (NF) similarity matrix $M^{F_s} \in \mathbb{R}^{L \times L}$, we fuse them to obtain an aggregating similarity matrix, where the value represents the similarity between the arbitrary two nodes $i$ and $j$ and can be calculated as follows:

$$M^{E_s}[i,j] = \frac{e_i \cdot e_j}{\|e_i\| \times \|e_j\|} \tag{2}$$

$$M^{F_s}[i,j] = \frac{f^{N_i} \cdot f^{N_j}}{\|f^{N_i}\| \times \|f^{N_j}\|} \tag{3}$$

$$M^{A_s} = M^{E_s} \circ \boldsymbol{W^{E_s}} + M^{F_s} \circ \boldsymbol{W^{F_s}} \tag{4}$$

where ∘ denotes Hadamard product between two matrixes. $\boldsymbol{W^{E_s}} \in \mathbb{R}^{L \times L}$ and $\boldsymbol{W^{F_s}} \in \mathbb{R}^{L \times L}$ are learnable weight matrixes, which weigh the importance of different dimensions of the different similarity matrixes. In $M^{A_s}$, when the correlation coefficient is larger than a certain threshold, we consider that there exists a connected relation between nodes; otherwise, the connected relation does not exist. To obtain the optimal threshold, we define a learnable parameter $\tau \in \mathbb{R}$ for automatic choice, and obtain the adjacent matrix of aggregating similarity graph through learning, which is denoted as:

$$A_{ij} = \begin{cases} 1 & for\ M^{A_s}[i,j] \geq \tau \\ 0 & for\ M^{A_s}[i,j] < \tau \end{cases} \tag{5}$$

In the heterogeneous dynamic graph, two objects can be connected through different semantic paths, which is called meta-path. However, the selection of meta-path has a strong subjective meaning, which is difficult for complex MTS. Therefore, we propose a classifying-based fixed-length sampled method to replace meta-path for extracting heterogeneous relation subgraph. Specifically, we divide the aggregating similarity graph into the corresponding classifying subgraphs, include discrete feature subgraph (DFS) $\mathbb{G}_{S^D(t)}$, continuous feature subgraph (CFS) $\mathbb{G}_{S^C(t)}$ and hybrid feature subgraph (HFS) $\mathbb{G}_{S^{CD}(t)}$ according to data types. We further make a random mask operation for the neighboring matrix of the subgraph and obtain the final neighboring matrix with different relations. The transformed heterogeneous graph structure is $A' = \{A^D, A^C, A^{CD}\}$. The random mask is conducive to exchange information between different similarity matrixes in the graph structure learning process, thus improving the accuracy of subsequent tasks and relieving the overfitting problem.

### 3.5. Graph representation learning for MTS

It can be seen from the learned heterogeneous graph structure that each type subgraph contains different semantic properties. Hence, to aggregate the node information from different types, we introduce a graph attention-based node embedding network and an attention-based channel aggregating network to construct the HFN for MTS. The structure is shown in **Figure 3**. Specifically, the obtained three subgraphs $A^D, A^C, A^{CD}$ learned by graph structure learning are inputted into three independent graph attention networks, to learn the importance of different types of nodes for the neighbors in the subgraphs. Moreover, the important neighboring information is aggregated to generate a new node embedding. As shown in **Figure 3**, taking the continuous numerical variable channel as an example, for the arbitrary node $v_i^C$ and its neighboring node $v_j^C$ in subgraph $A^C$, we perform self-attention in the nodes. The attention coefficient representing their relation importance can be calculated as:

$$\xi_{ij} = att(\boldsymbol{W}f^{C_i}, \boldsymbol{W}f^{C_j}; A^C) \tag{6}$$

where $f^{C_i} \in \mathbb{R}^{1 \times \omega'}$ and $f^{C_j} \in \mathbb{R}^{1 \times \omega'}$ are mapped node feature vectors, $\boldsymbol{W} \in \mathbb{R}^{\omega'' \times \omega'}$ is shared weight matrix. $\omega'$ and $\omega''$ are the calculated node feature vector dimensions before and after the embedding. After obtaining the importance of subgraph-based node pairs, we normalize them via the SoftMax function and obtain weight coefficient $\alpha_{ij}$:

$$\alpha_{ij} = softmax(\sigma_{ij}) = \frac{exp\left(\delta(\vec{a}^T[\boldsymbol{W}f^{C_i}||\boldsymbol{W}f^{C_j}])\right)}{\sum_{\eta \in N_i^C} exp\left(\delta(\vec{a}^T[\boldsymbol{W}f^{C_i}||\boldsymbol{W}f^{C_\eta}])\right)} \tag{7}$$

where $\delta$ is the activation function, and LeakyReLU function is usually adopted [54]. $\vec{a} \in \mathbb{R}^{2\omega''}$ is the learnable weight vector, which denotes the information concatenation of the two nodes. Finally, the output of each node can be obtained through aggregating its neighboring nodes. Multi-head attention mechanism is proven to be beneficial in the learning process of stabilizing self-attention [59]. To be convenient for training, we perform an average operation to aggregate the results handled by multi-head attention. After the graph attention-based nodes embed into the network, the implicit vector can be represented as:

$$h_i'^C = \sigma\left(\frac{1}{H}\sum_{h=1}^{H}\sum_{j\in N_i^C}\alpha_{ij}^h W^h f^{C_j}\right) \quad (8)$$

where $N_i^C$ is the set of nodes $i$'s neighbors in the continuous subgraph. $H$ denotes the number of multi-head attention mechanism head. According to the same computation method, we can obtain the node implicit vectors of discrete subgraph and mixed subgraph represented by $h_i'^D$ and $h_i'^{CD}$.

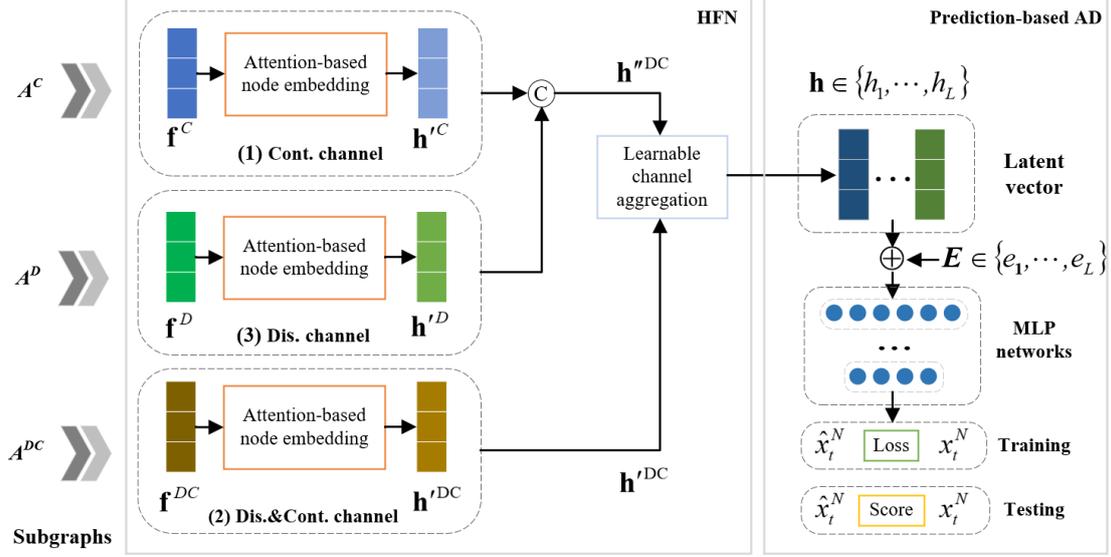

Figure 3 Heterogeneous feature network structure of MTS

To address the node semantic importance of different types in the heterogeneous graph, we put forward an attention-based multi-channel node embedding aggregating network. We can clearly see from **Figure 3** that the node implicit vectors $\boldsymbol{h}'^C$ and $\boldsymbol{h}'^D$ singly from a continuous channel and discrete channel first are concatenated in feature dimension to obtain the global node implicit vector $\boldsymbol{h}''^{DC}$. The main purpose is to achieve the joint embedding representation learning of all nodes simultaneously. Then $\boldsymbol{h}''^{DC}$ and the node implicit vector $\boldsymbol{h}'^{CD}$ from the mixed channel are sent to the multi-channel node embedding aggregating network for aggregating their heterogeneous information. The aggregating network automatically learns the importance degree $\beta$ of the embedding vectors between different channel node implicit vectors, which can be explained as the contribution of the node correlation due to the different types of variables. The final embedding vector is computed as follows:

$$\boldsymbol{h} = \beta(\boldsymbol{h}'^C\|\boldsymbol{h}'^D) + (1-\beta)\boldsymbol{h}'^{CD} \quad (9)$$

where $\beta \in \mathbb{R}$ is learnable parameter representing the importance degree of the embedding vectors between different channel node implicit vectors, and $\|$ is a concatenation operation.

### 3.6. Prediction-based anomaly detection pipeline

From the above node heterogeneous feature learning network, we obtain new embedding representations of all nodes. Finally, as shown in **Figure 3**, we input the embedding data fused with $\boldsymbol{h}$ and embedding vector $\boldsymbol{E}$ into the MLP layer to have the predicted value $\hat{x}_t^N$ of all sensors at the moment $t$:

$$\hat{x}_t^N = SeLU\big(f(\boldsymbol{h}\oplus \boldsymbol{E})\big) \quad (10)$$

where $f(\cdot)$ is multiple layers of MLP output layer. $SeLU$ is activation function, and $\oplus$ is addition operation.

At the training stage, we adopt MSE as the loss function of the model:

$$\mathcal{L}_{mse} = \frac{1}{L}\sum_i^L (x_t^N - \hat{x}_t^N)^2 \qquad (11)$$

After the training is finished, we apply the network to perform real-time abnormal detection tasks. By comparing the predicted and original values of the input, we calculate the condition scores of each sample in time-series data. We define the difference between the original value and predicted value as the condition scores. To eliminate the effect of different variable dimensions, we normalize the condition scores. Finally, the condition score is computed as follows:

$$Score_i = \frac{|x_t^{N_i} - \hat{x}_t^{N_i}| - IQR_i}{\mu_i + 1} \qquad (12)$$

where $IQR_i$ denotes an interquartile range of the predicted value of the $i$th variable, $\mu_i$ is its median. To achieve the anomaly positioning, we take the largest value of $Score_i$ as the condition score of overall record data at the moment $t$, as denoted by $Score = max(Score_i)$. Finally, if the $Score$ is larger than the threshold, this record is judged as an anomaly. However, because the threshold selection refers to complicated domain knowledge and the selection methods are various depending on the applications [60], this paper will not further explore the selection method for the threshold. The experiment in the subsequent section will report the optimal value of each evaluating metric (see details in Section 4.3).

## 4. Experiments

We employ extensive experiments on two open and one private real-world datasets to answer the following research questions:
(1) **Model performances**
   **RQ1**: Whether the proposed model is more optimal than the baseline models?
   **RQ2**: How each component of the model affects the model?
(2) **Abnormal detection**
   **RQ3**: How the proposed approach detects anomalies?
   **RQ4**: How the detection results locate anomalies?

### 4.1. Benchmark datasets

The selected three datasets contain two datasets (SWaT and WADI) based on water treatment simulator testbed and a real-world dataset from a large-scale wind farm (WTD). The statistical data of the datasets are given in Table 1:

**Table 1 Statistics of the datasets**

| Items | SWaT | WADI | WTD |
|---|---|---|---|
| Time series (C/D) | 51 (25/26) | 123 (68/55) | 37(31/6) |
| Training dataset | 496800 | 784571 | 1000000 |
| Testing dataset | 449919 | 172803 | 940000 |
| Anomaly Rate (%) | 11.97% | 5.99% | 20.64% |

**Secure Water Treatment (SWaT) Dataset** [61]. This dataset was collected from a six-stage Secure Water Treatment (SWaT) testbed. SWaT represents a scaled-down version of a real-world industrial water treatment plant. It took 11 days for the data collection process, which run with normal operation mode during the first seven days, and constituted a training dataset. During the later four days, the testbed was implemented by intermittent network and physical attacks, which constituted the labeled testing dataset. The data were collected once every second, containing 51 time-series features, including 25 continuous features and 26 discrete categorical features. We chose this dataset for case study, and the primary sensors or actuators involved are shown in the **Table 2** below.

**Table 2 Primary sensors or actuators for case study**

| No. | Name | Type | Description |
|---|---|---|---|
| 1 | FIT-401 | Sensor (continuous) | Flow transmitter to control the UV dechlorinator. |
| 2 | UV-401 | Actuator (discrete) | Dechlorinator to remove the chlorine from water. |
| 3 | FIT-504 | Sensor (continuous) | Flow meter, a RO re-circulation flow meter. |
| 4 | P-501 | Actuator (discrete) | Pump to pump the dechlorinated water to RO. |
| 5 | LIT-401 | Sensor (continuous) | Level transmitter to regulate the RO feed water tank level. |
| 6 | LIT-101 | Sensor (continuous) | Level transmitter to regulate the raw water tank level. |
| 7 | FIT-601 | Sensor (continuous) | Flow meter a UF backwash flow meter. |
| 8 | AIT-504 | Sensor (continuous) | RO permeate conductivity analyzer to measure the NaCl level. |
| 9 | AIT-201 | Sensor (continuous) | Conductivity analyzer to measure the NaCl level. |

**Water Distribution (WADI) Dataset** [62]. This dataset was collected from a water distribution testbed (WADI). It took 16 days for the data collection process. During the last two days, the attack was launched to the testbed with different intentions and time intervals, and the duration of the attack lasted between 1.5 to 30 minutes to acquire the abnormal operating data. The data were collected once every second, containing 123 time-series features, including 68 continuous features and 55 categorical features.

**Wind Turbine Dataset (WTD).** This dataset was collected from a large-scale wind farm [63]. It lasted 1 to 2 years for the data collection process. At the training stage, there are no abnormal operating data since only the time-based maintenance process was arranged for the wind turbines, while at the testing stage, the abnormal operating data were detected in the repairing process. All data have been labeled by the experts. The data were collected once every 10 minutes, containing 37 time-series features, including 31 continuous features and 6 categorical features.

### 4.2. Baseline models

We first compare the FHN model with the most advanced approaches including LSTM-VAE [64], USAD [65], MAD-GAN [17], graph network based MTAD-GAT [11] and GDN [53]. These approaches extensively concern with the cross-time and cross-sequence correlation of MTS. The approaches based on sequence reconstruction or prediction are used to learn the representations of the whole time series. Moreover, the anomalies are judged by the reconstructing or predicting errors.

Furthermore, we compare the proposed approach with those classic shallow anomaly detection approaches, including PCA [66], Isolation Forest [67] and LightGBM [68]. These shallow detection methods are regarded as the relatively direct abnormal detection methods, which usually can directly locate the outlier. Moreover, to complete the anomaly detection in temporally related contexts has also attracted the interests for the researchers, such as LSTM-NDT [69]. The idea underlying this method is to model the temporal features of the data, predict the corresponding values, and then judge whether the anomalies occur by comparing the deviation between the real value and the predicted value.

### 4.3. Evaluation

(1) Metrics

We select precision, recall and F1 as the evaluating metrics of the model, where $Precision = \frac{TP}{TP+FP}$, $Recall = \frac{TP}{TP+FN}$, $F1 = \frac{2 \times Precision \times Recall}{Precision+Recall}$. $TP$, $FP$ and $FN$ refer to true positives, false positives, and false negatives, respectively. These metrics are required to be obtained with a certain threshold. Hence, due to the different threshold selection methods among different tasks, there exist large differences in the metric values. Therefore, to avoid introducing additional hyperparameters, we report the evaluation metrics values when the optimal F1 value is obtained. The threshold value is determined by traversing between the maximum and the minimum scores of the testing dataset.

We calculate the condition scores that decide the abnormal degree of the overall dataset based on eq. (12). Noted that in unsupervised anomaly detection for MTS (USAD) [65] and temporal hierarchical one-class network (THOC) [70], the authors applied a specific evaluation method, called point adjust, making F1 value higher and close to 1. It has been proved that the capability of the model may be highly evaluated [71]. Hence, for the comparison, we apply the open-source code of USAD, and utilize the same parameters of the model in this paper to calculate the performance metrics without adjustment.

**(2) Setup**

We use Pytorch to achieve the HFN and its variants. Moreover, the model is trained on a server with Intel(R) Xeon(R) Gold 5218R CPU @ 2.1GHz and NVIDIA GeForce RTX 3090 graphics cards. We select Adam optimizer to train the model. Meanwhile, we adopt early stopping to relieve overfitting. The maximum training epoch is set to be 100. If the loss is less than 0.0001 after 10 epochs, the training stops automatically and the optimal model is saved. Furthermore, we set the sliding window length to be 15 for the input data sequence, the dimension after node embedding and feature projection to be 64, and the dimension of the final embedding vector to be 10.

### 4.4. Evaluation results

Table 3 Precision, recall and F1 values of HFN and all baseline methods on different datasets.

| model | SWaT | | | WADI | | | WTD | | |
|---|---|---|---|---|---|---|---|---|---|
| | Prec | Rec | F1 | Prec | Rec | F1 | Prec | Rec | F1 |
| PCA | 0.249 | 0.216 | 0.230 | 0.395 | 0.056 | 0.100 | 0.160 | 0.513 | 0.244 |
| Isolation Forest | 0.951 | 0.588 | 0.727 | 0.299 | 0.158 | 0.207 | 0.278 | 0.953 | 0.430 |
| LightGBM | 0.783 | 0.666 | 0.719 | 0.989 | 0.153 | 0.270 | 0.237 | 0.602 | 0.340 |
| LSTM-NDT | 0.982 | 0.688 | 0.809 | 0.758 | 0.328 | 0.457 | 0.365 | 0.736 | 0.497 |
| LSTM-VAE | 0.962 | 0.599 | 0.740 | 0.878 | 0.145 | 0.250 | 0.165 | 0.550 | 0.254 |
| DAGMM | 0.470 | 0.666 | 0.551 | 0.544 | 0.267 | 0.360 | 0.164 | 0.242 | 0.195 |
| OmniAnomaly | 0.983 | 0.650 | 0.782 | 0.995 | 0.130 | 0.230 | - | - | - |
| USAD | 0.985 | 0.661 | 0.792 | 0.995 | 0.132 | 0.233 | 0.157 | 0.417 | 0.228 |
| MAD-GAN | 0.990 | 0.637 | 0.770 | 0.414 | 0.339 | 0.370 | - | - | - |
| MTAD-GAT | 0.991 | 0.633 | 0.772 | 0.988 | 0.153 | 0.265 | 0.128 | 0.397 | 0.193 |
| GDN | 0.994 | 0.681 | 0.810 | 0.975 | 0.402 | 0.570 | 0.385 | 0.937 | 0.546 |
| **HFN** | 0.973 | **0.758** | **0.852** | 0.827 | **0.413** | 0.551 | **0.505** | 0.837 | **0.630** |

The optimal metric values are shown in bold in Table 2. For the datasets SWaT and WADI, we refer to the results in USAD [65] and graph deviation network (GDN) [53]. For WTD dataset, to guarantee the objectivity of the results, we only report the metrics from the obtained open code approaches.

**RQ1:Whether the proposed model is more optimal than the baseline model?**

We can observe from **Table 2** that HFN shows a good abnormal detection capability with remarkable performance improvements on SWaT and WTD. The improvement range of the proposed approach is 5% to 14%, as compared with the optimal baseline models. The optimal baseline GDN outperforms our approach in terms of F1; however, our approach has a more optimal recall rate. It is acceptable in real scenarios because we hope to detect more anomalies. In short, HFN outperforms the selected baselines in terms of the overall performances, because it not only concerns with the traditional spatial-temporal correlation, but also obtains its heterogeneous attributes from different types of data, making the model more robust. Moreover, we observe that prediction-based algorithms such as HFN, GDN and LSTM-NDT outperform the reconstruction-based algorithms such as LSTM-VAE and USAD on these datasets, indicating that the prediction-based models have an advantage in the streaming abnormal detection tasks with a single- timestamp value as the target. The temporal information is also very vital in the tasks for MTS abnormal detection. The results of LSTM-NDT show that HFN outperforms all baselines except GDN. The PCA result is dissatisfactory, because it gives more attentions to the point anomalies without spatial-temporal correlation being considered.

Specifically, among these abnormal detection approaches, GDN, MTAD-GAT and HFN adopt the graph attention network to capture the temporal and feature correlations. Therefore, these types of models achieve good results on all datasets. GDN approach recodes multidimensional data at each moment, and utilizes its strong structural learning capability of graph attention network to learn coupling relations between different sensors. However, it does not consider the heterogeneity of data. MTAD-GAT approach also captures time-dimension information through an attention mechanism. Although it considers the spatial-temporal correlation of MTS, it requires a configuration of hyper-parameters for fusing the prediction-based and reconstruction-based condition scores, leading to the evident differences in results when this approach is applied to different datasets.

**RQ2: How each component of the model affects the model?**

We utilize SWaT and WADI datasets to study the necessity of five components of our approach, namely, node embedding similarity matrix (NE), node feature similarity matrix (NF), discrete feature subgraph (DFS), continuous feature subgraph (CFS), and hybrid feature subgraph (HFS). As shown in **Figure 4**, we successively exclude the corresponding component from the experiments to observe its effect on the model performance. The key idea of our approach is to learn the potential steady representations from heterogeneous MTS. Hence, first, we exclude NE or NF to study whether the heterogeneous information is learned. Second, we discuss the anomaly detection performance when we only use HFS or DFS and CFS. Specifically:

(1) Excluding NF (expressed as "-NF") degrades the overall performance of the approach and has a great influence on WADI dataset. This indicates that NF is in favor of feature extraction with high-dimensional dataset for the model; however, NF is not the key factor to determine the model performance.

(2) Excluding NE (expressed as "-NE") degrades the performances clearly, which implies that NE has an evident advantage in the graph structure learning process.

(3) Excluding DFS and CFS (expressed as "-DFS" and "-CFS") degrades the model performance; however, the descend range of model performance is less than that of NE. This approach is actually degenerated to the processing of isomorphic graphs, leading to the loss of heterogeneous information.

(4) Excluding HFS (expressed as "-HFS") degrades the model performance; however, it is superior to the cases when DFS and CFS are totally excluded. This indicates that the interaction between different types of sensors in the hybrid subgraph plays a complementary role in extracting the follow-up HFN heterogeneous information.

To sum up, it is necessary to extract heterogeneous structure information in the MTS datasets. The heterogeneous information can present different weights in the model according to the attention mechanism, which helps to improve the abnormal detection performance.

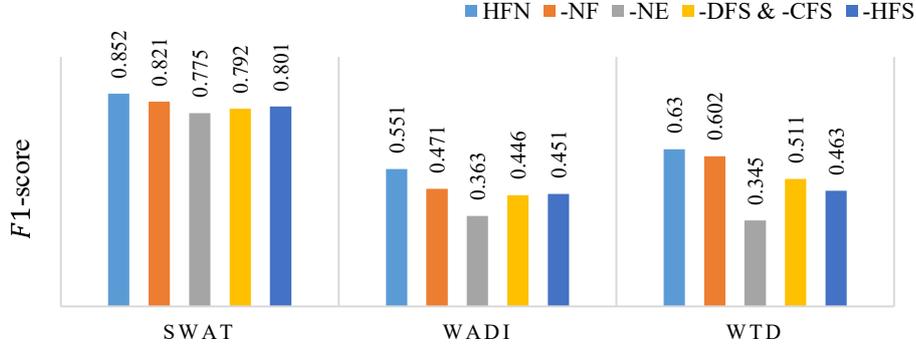

**Figure 4 Effects of different HFN components on anomaly detection performance**

### 4.5. Anomaly interpretation

**RQ3: How the proposed approach detects anomalies?**

**Figure 5** shows the abnormal detection results on SWaT testing dataset, where **Figure 5**(a) represents the actual data anomalies on this dataset, including network and physical attacks directed at the Secure Water Treatment (SWaT) testbed within the continuous four days. The data are labeled as 1 if the system is attacked at a certain timestamp; otherwise, it is labeled as 0. **Figure 5**(b) represents the results of HFN anomaly detection, where the orange shadow represents the detected anomalies, the blue curve represents the condition scores calculated as described in Section 3.6, and the red straight line represents the threshold when the optimal F1 is obtained on the testing dataset. It can be seen from **Figure 5**(b) that, aside from a few anomalies that are very difficult to distinguish possibly due to labeling errors, our approach accurately identifies the most anomalies. According to the instructions provided by SWAT dataset [61], we select an attack case to further interpret the abnormal detection capability of HFN. As shown in **Figure 5**(a), the attack starts from 14:16:00, 28/12/2015 to 14:28:00, 28/12/2015 against FIT401, UV401 and P501, where FIT401 is the flow transmitter for measuring the flow of UV de-chlorinator, UV401 is de-chlorinator for removing chlorine from water, and P501 is pump actuator for pumping the dechlorinated water to reverse osmosis. During the attack, as shown in Figure 6, the flow value (continuous value) of FIT401 is set twice to the value deviating from the normal mode. Meanwhile, the actuators UV401 and P501 (discrete value), which should be kept to an open state, are forcefully closed.

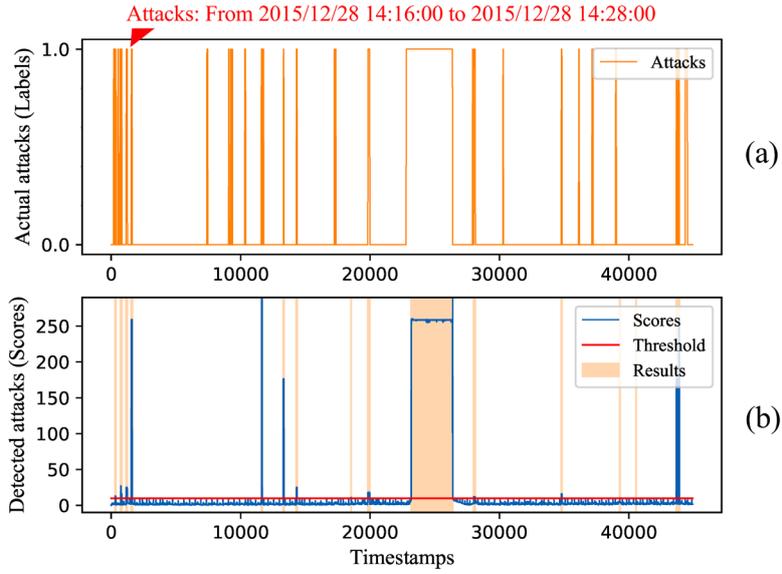

**Figure 5 SWaT dataset anomaly detection results**

**Figure 6** shows the curves of actual and predicted values of attack-related sensors and actuators and the HFN anomaly detection results. In order to reduce the influence of data dimensions and accelerate the convergence of the model, we have standardized the values of the dataset by min-max normalization. It is worth noting that we used the same normalized parameters for both the training dataset and the testing dataset, which is why the normalized data of the testing data shown in **Figure 6** has negative values. This was done to reduce the impact of testing dataset information leakage on the model performance. In the real water treatment process, the unit of the flow sensors values are gallons per minute (GPM), while the actuators have two conditions: 0 means turn on and - 1 means turn off.

It can be seen from **Figure 6**(a), (c) and (d) that before the attack, the predicted values of HFN are consistent with the actual values, where the prediction for both continuous variables and discrete variables achieves good results. In the attack process, the flow variation arises from the prediction result of FIT401 and UV401 simultaneously. This is due to the interaction among these variables in the actual water treatment system. A larger deviation between the predicted value and the actual value would provide a better basis for abnormal detection. Note that although the experiment personnel did not launch the attack on FIT504 sensor in the attack process, we can see from **Figure 6**(b) and (d) that the value changes of FIT504 are still detected, which is due to being abnormally closed caused by the attack on P501. We can observe from the detection results in **Figure 6**(e) that the proposed approach shows a good detection capability of such complex anomalies. These anomalies have been resulted from attacks to different types of sensors, including continuity, discreteness and their correlation, which represent real scenarios.

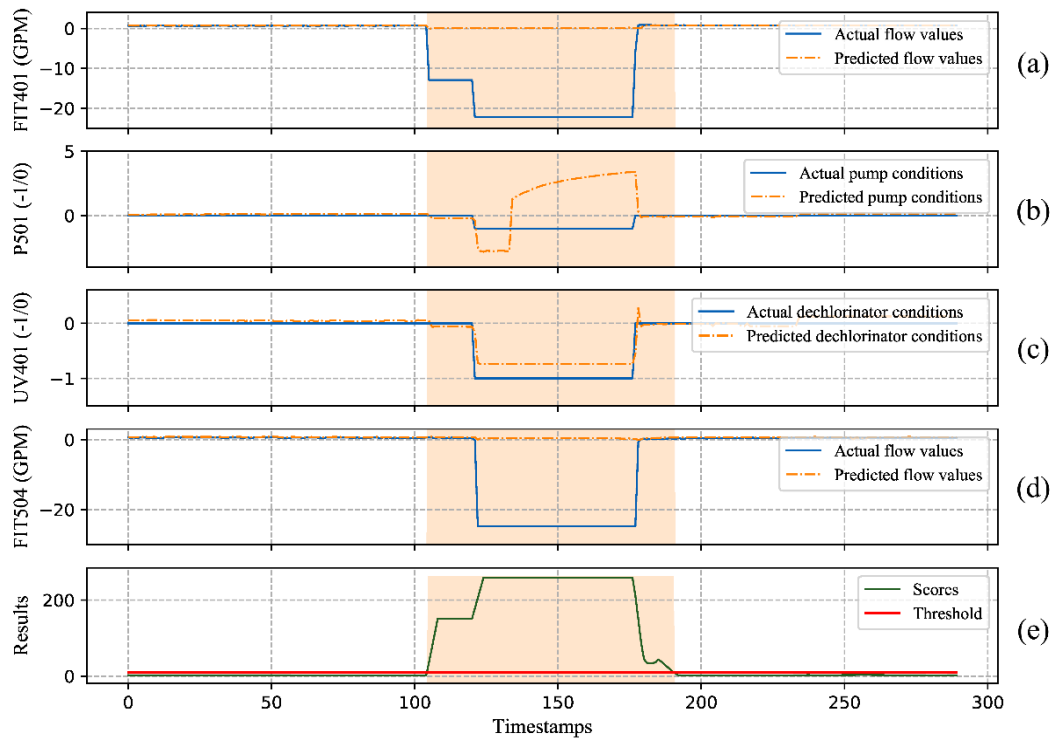

**Figure 6** Abnormal detection case. The orange shadow represents the detected anomalies. The blue curve denotes the actual value of sensor or actuator. The orange dotted line represents the predicted value. The green and red curves in Fig. 6(e) represent condition score and the threshold of the optimal F1, respectively.

**RQ4：How the detection results locate anomalies?**

From the above analysis, we can see that our method can successfully detect the occurrence of anomalies. However, we cannot assume that all the variables in a real complicated system are of the same significance. In other words, the variables associated with a particular system component will be impacted to varying degrees of operation when that component is attacked or behaves abnormally. Therefore, it is necessary to locate variables that have been strongly impacted by the attack, thus helping system maintenance personnel to rapidly find and solve the problems. We use the prediction error of each time-series sensor to represent the condition score of the sequence where the sensor with the maximum score times is considered to have the possibility of the biggest anomalies. **Figure 7** shows the number of times when the condition scores are above the threshold for different sensors within the attack period in the case analysis. It can be known from the figure that the sensors FIT504 and FIT401 have the maximum score times, which is consistent with the attacks where the experiment personnel made to the sensor FIT401 and pump actuator P501 during the tests. The turn-off attacks on P501 caused a sharp drop in the FIT504 flow values, as shown in Figure 6(d), since they are physically connected. On the contrary, we can also speculate which component of the system has been attacked or abnormal according to the maximum score times. In this case, during real operation and maintenance, particular attentions should be paid to and checks should be made on the locations relating to FIT504, FIT401, and LIT401.

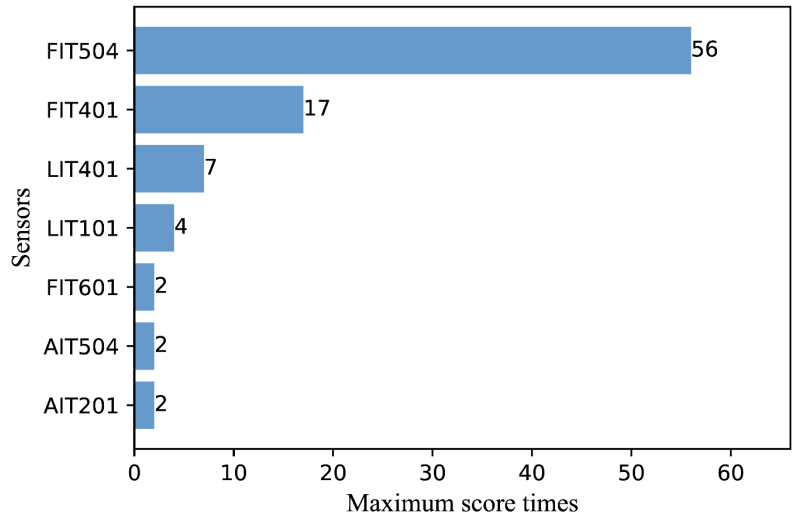

Figure 7 Maximum number of sensor scores in the abnormal dataset

### 4.6. Model discussion

To further illustrate how the heterogeneous relation in time series is learned and takes effect on the abnormal detection, we explain it through the similarity matrix before and after the anomalies due to the attacks. **Figure 8** and **Figure 9** represent different similarity matrices before and after the attack on SWaT, respectively. Its similarity value range is [- 1,1], and the closer to 1, the stronger the similarity is. Overall, HFN aggregates the similarities of sensor signals from different perspectives to represent its heterogeneous information. Embedding similarity matrix learns the structural information among different sensors globally from the training data. Hence, similar features are shown in Figure 8(b) and Figure 9(b) under abnormal and normal states. However, concerning the feature similarity, we can see clearly that there exist significant differences in feature similarity between Figure 8(a) and Figure 9(a) at different timestamps, because the data vary with time. Ignoring this part of information always degrades the abnormal detection performance.

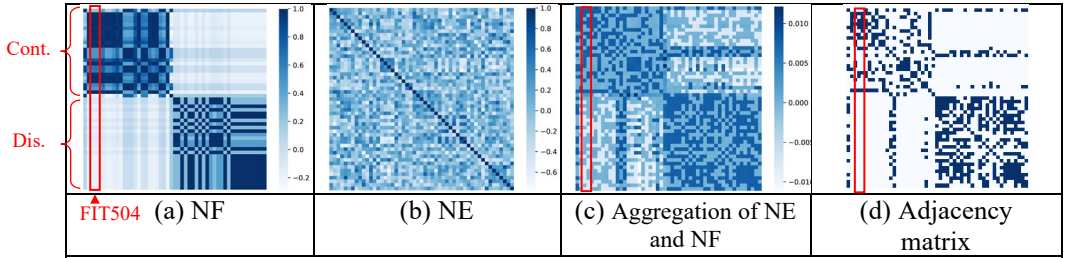

Figure 8 Example of similarity subgraphs under normal conditions

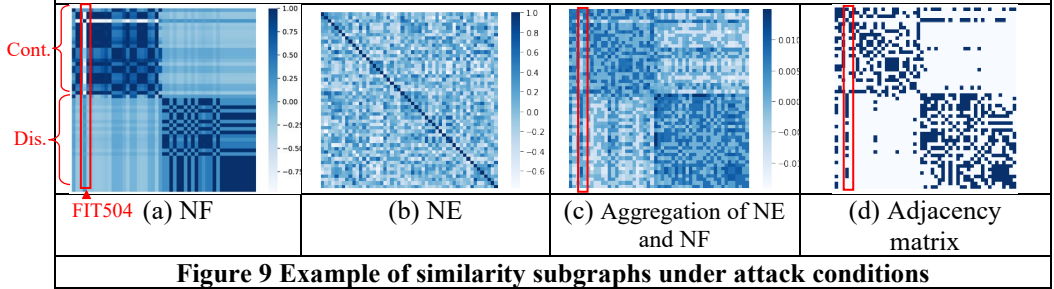

Figure 9 Example of similarity subgraphs under attack conditions

Specifically, as shown in **Figure 8**(a), before the attacks on FIT401, UV401 and P501, the similarity values of FIT504 flow value and other continuous variable sensor values are close to 1. However, after the attack, we can see from **Figure 9**(a) that the similarity value varies to -0.75. The sudden change indicates that the sensor anomalies occur, while there are slight variations in the embedding similarity. After comparing the adjacent matrix before and after the attack in **Figure 8**(d) and **Figure 9**(d), we can find that the changes in feature similarity cause the changes in the connection relation to improve the ability of the algorithm in capturing dynamic feature correlation. This further demonstrates that HFN, by aggregating the global data learning-based embedding similarity matrix and the feature similarity matrix at a specific timestamp, can better capture the normal and abnormal conditions in MTS.

## 5. Conclusions

In this paper, we propose a novel heterogeneous feature network for MTS anomaly detection. This approach is able to learn the complex heterogeneous structural information and temporal information between MTS data. Therefore, it is suitable for abnormal detection in real scenarios where the dataset comprises continuous numerical variables and discrete categorical variables simultaneously. The extensive experiments indicate that our approach outperforms the baseline models by assessing two open datasets from water treatment plants and a private dataset from a wind power plant. Furthermore, our approach demonstrates a good abnormal interpretability and can help operation and maintenance personnel rapidly discover and locate the anomalies. In the future, we can continue to improve the proposed algorithm with more real and complicated heterogeneous datasets including the combined time series data and text data, thus further improving the accuracy and practicability of the approach.

# Acknowledgment

The work is supported by National Natural Science Foundation of China (62006236), NUDT Research Project (ZK20-10), National Key Research and Development Program of China (2020YFA0709803), Hunan Provincial Natural Science Foundation (2020JJ5673), National Science Foundation of China (U1811462), National Key R&D project by Ministry of Science and Technology of China (2018YFB1003203), and Autonomous Project of HPCL (201901-11, 202101-15).

# 6. References

1. Myers, D., Suriadi, S., Radke, K. and Foo, E., Anomaly detection for industrial control systems using process mining. *Computers & Security* **2018,** *78*, 103-125.
2. Bhamare, D., Zolanvari, M., Erbad, A., Jain, R., Khan, K. and Meskin, N., Cybersecurity for industrial control systems: A survey. *computers & security* **2020,** *89*, 101677.
3. Das, T. K., Adepu, S. and Zhou, J., Anomaly detection in industrial control systems using logical analysis of data. *Computers & Security* **2020,** *96*, 101935.
4. Almalawi, A., Yu, X., Tari, Z., Fahad, A. and Khalil, I., An unsupervised anomaly-based detection approach for integrity attacks on SCADA systems. *Computers & Security* **2014,** *46*, 94-110.
5. Zhan, J., Wang, R., Yi, L., Wang, Y. and Xie, Z., Health assessment methods for wind turbines based on power prediction and mahalanobis distance. *International Journal of Pattern Recognition and Artificial Intelligence* **2019,** *33* (02), 1951001.
6. Pang, G., Shen, C., Cao, L. and Hengel, A. V. D., Deep learning for anomaly detection: A review. *ACM Computing Surveys (CSUR)* **2021,** *54* (2), 1-38.
7. Wu, P. and Liu, J., Learning causal temporal relation and feature discrimination for anomaly detection. *IEEE Transactions on Image Processing* **2021,** *30*, 3513-3527.
8. Lindemann, B., Maschler, B., Sahlab, N. and Weyrich, M., A survey on anomaly detection for technical systems using LSTM networks. *Computers in Industry* **2021,** *131*, 103498.
9. Bai, S., Kolter, J. Z. and Koltun, V., An Empirical Evaluation of Generic Convolutional and Recurrent Networks for Sequence Modeling. *arXiv preprint arXiv:1803.01271.* **2018**.
10. Blázquez-García, A., Conde, A., Mori, U. and Lozano, J. A., A review on outlier/anomaly detection in time series data. *ACM Computing Surveys (CSUR)* **2021,** *54* (3), 1-33.
11. Zhao, H., Wang, Y., Duan, J., Huang, C. and Zhang, Q., Multivariate Time-series Anomaly Detection via Graph Attention Network. **2020**.
12. Du, S., Li, T., Yang, Y. and Horng, S.-J., Multivariate time series forecasting via attention-based encoder–decoder framework. *Neurocomputing* **2020,** *388*, 269-279.
13. Kim, T.-Y. and Cho, S.-B., Predicting residential energy consumption using CNN-LSTM neural networks. *Energy* **2019,** *182*, 72-81.
14. He, K., Zhang, X., Ren, S. and Sun, J. In *Deep Residual Learning for Image Recognition*, IEEE Conference on Computer Vision & Pattern Recognition, 2016.
15. Wu, Z., Pan, S., Long, G., Jiang, J., Chang, X. and Zhang, C. In *Connecting the dots: Multivariate time series forecasting with graph neural networks*, Proceedings of the 26th ACM SIGKDD International Conference on Knowledge Discovery & Data Mining, 2020; pp 753-763.
16. Guo, Y., Liao, W., Wang, Q., Yu, L., Ji, T. and Li, P. In *Multidimensional time series anomaly*

*detection: A gru-based gaussian mixture variational autoencoder approach*, Asian Conference on Machine Learning, PMLR: 2018; pp 97-112.

17. Li, D., Chen, D., Shi, L., Jin, B., Goh, J. and Ng, S. K., MAD-GAN: Multivariate Anomaly Detection for Time Series Data with Generative Adversarial Networks. **2019**.

18. Borisov, V., Leemann, T., Seßler, K., Haug, J., Pawelczyk, M. and Kasneci, G., Deep neural networks and tabular data: A survey. *arXiv preprint arXiv:2110.01889* **2021**.

19. Nazabal, A., Olmos, P. M., Ghahramani, Z. and Valera, I., Handling incomplete heterogeneous data using vaes. *Pattern Recognition* **2020,** *107*, 107501.

20. Sun, Y., Han, J., Yan, X., Yu, P. S. and Wu, T., Pathsim: Meta path-based top-k similarity search in heterogeneous information networks. *Proceedings of the VLDB Endowment* **2011,** *4* (11), 992-1003.

21. Wang, X., Ji, H., Shi, C., Wang, B., Ye, Y., Cui, P. and Yu, P. S. In *Heterogeneous graph attention network*, The world wide web conference, 2019; pp 2022-2032.

22. Lngkvist, M., Karlsson, L. and Loutfi, A., A review of unsupervised feature learning and deep learning for time-series modeling. *Pattern Recognition Letters* **2014,** *42*, 11-24.

23. Li, J., Izakian, H., Pedrycz, W. and Jamal, I., Clustering-based anomaly detection in multivariate time series data. *Applied Soft Computing* **2020,** *100* (4), 106919.

24. Liu, X., Li, M., Qin, S., Ma, X. and Wang, W., A Predictive Fault Diagnose Method of Wind Turbine Based on K-Means Clustering and Neural Networks. *網際網路技術學刊* **2016**.

25. Izakian, H. and Pedrycz, W. In *Anomaly detection in time series data using a fuzzy c-means clustering*, 2013 Joint IFSA world congress and NAFIPS annual meeting (IFSA/NAFIPS), IEEE: 2013; pp 1513-1518.

26. Jones, M., Nikovski, D., Imamura, M. and Hirata, T. In *Anomaly detection in real-valued multidimensional time series*, International Conference on Bigdata/Socialcom/Cybersecurity. Stanford University, ASE. Citeseer, Citeseer: 2014.

27. Nascimento, E. G. S., de Lira Tavares, O. and De Souza, A. F. In *A cluster-based algorithm for anomaly detection in time series using Mahalanobis distance*, Proceedings on the International Conference on Artificial Intelligence (ICAI), The Steering Committee of The World Congress in Computer Science, Computer …: 2015; p 622.

28. Pu, G., Wang, L., Shen, J. and Dong, F., A hybrid unsupervised clustering-based anomaly detection method. *Tsinghua Science and Technology* **2020,** *26* (2), 146-153.

29. Jia, W., Shukla, R. M. and Sengupta, S. In *Anomaly Detection using Supervised Learning and Multiple Statistical Methods*, IEEE International Conference on Machine Learning and Applications, 2019.

30. Görnitz, N., Kloft, M., Rieck, K. and Brefeld, U., Toward supervised anomaly detection. *Journal of Artificial Intelligence Research* **2013,** *46*, 235-262.

31. Ruff, L., Vandermeulen, R. A., Görnitz, N., Binder, A., Müller, E., Müller, K.-R. and Kloft, M., Deep semi-supervised anomaly detection. *arXiv preprint arXiv:1906.02694* **2019**.

32. Villa-Pérez, M. E., Alvarez-Carmona, M. A., Loyola-González, O., Medina-Pérez, M. A., Velazco-Rossell, J. C. and Choo, K.-K. R., Semi-supervised anomaly detection algorithms: A comparative summary and future research directions. *Knowledge-Based Systems* **2021,** *218*, 106878.

33. Wang, R., Ma, X., Jiang, C., Ye, Y. and Zhang, Y., Heterogeneous information network-based music recommendation system in mobile networks. *Computer Communications* **2020,** *150*, 429-437.

34. Liang, X., Ma, Y., Cheng, G., Fan, C., Yang, Y. and Liu, Z., Meta-path-based heterogeneous graph neural networks in academic network. *International Journal of Machine Learning and Cybernetics* **2022,**

*13* (6), 1553-1569.

35. Deng, X., Long, F., Li, B., Cao, D. and Pan, Y., An influence model based on heterogeneous online social network for influence maximization. *IEEE Transactions on Network Science and Engineering* **2019,** *7* (2), 737-749.

36. Chen, X., Yin, J., Qu, J. and Huang, L., MDHGI: matrix decomposition and heterogeneous graph inference for miRNA-disease association prediction. *PLoS computational biology* **2018,** *14* (8), e1006418.

37. Sun, Y., Gao, J., Hong, X., Mishra, B. and Yin, B., Heterogeneous tensor decomposition for clustering via manifold optimization. *IEEE transactions on pattern analysis and machine intelligence* **2015,** *38* (3), 476-489.

38. Liu, Y., Luo, X. and Yang, X. In *Semantics and structure based recommendation of similar legal cases*, 2019 IEEE 14th International Conference on Intelligent Systems and Knowledge Engineering (ISKE), IEEE: 2019; pp 388-395.

39. Wang, C., Chi, C. H., Wei, Z. and Wong, R., Coupled Interdependent Attribute Analysis on Mixed Data. **2015**.

40. Jian, S., Cao, L., Pang, G., Kai, L. and Hang, G., 17IJCAI Embedding -based Representation of Categorical Data by Hierarchical Value Coupling Learning. **2017**.

41. Cai, H., Zheng, V. W. and Chang, K. C.-C., A comprehensive survey of graph embedding: Problems, techniques, and applications. *IEEE Transactions on Knowledge and Data Engineering* **2018,** *30* (9), 1616-1637.

42. Zhao, J., Wang, X., Shi, C., Hu, B., Song, G. and Ye, Y. In *Heterogeneous graph structure learning for graph neural networks*, 35th AAAI Conference on Artificial Intelligence (AAAI), 2021.

43. Fu, X., Zhang, J., Meng, Z. and King, I. In *Magnn: Metapath aggregated graph neural network for heterogeneous graph embedding*, Proceedings of The Web Conference 2020, 2020; pp 2331-2341.

44. Wang, X., Liu, N., Han, H. and Shi, C. In *Self-supervised heterogeneous graph neural network with co-contrastive learning*, Proceedings of the 27th ACM SIGKDD Conference on Knowledge Discovery & Data Mining, 2021; pp 1726-1736.

45. Hu, Z., Dong, Y., Wang, K. and Sun, Y. In *Heterogeneous graph transformer*, Proceedings of The Web Conference 2020, 2020; pp 2704-2710.

46. Yang, L., Xiao, Z., Jiang, W., Wei, Y., Hu, Y. and Wang, H. In *Dynamic Heterogeneous Graph Embedding Using Hierarchical Attentions*, Springer, Cham, 2020.

47. Bloomfield, P., *Fourier analysis of time series: an introduction*. John Wiley & Sons: 2004.

48. Shwartz-Ziv, R. and Armon, A., Tabular data: Deep learning is not all you need. *Information Fusion* **2022,** *81*, 84-90.

49. Zhu, Y., Xu, W., Zhang, J., Liu, Q., Wu, S. and Wang, L., Deep graph structure learning for robust representations: A survey. *arXiv preprint arXiv:2103.03036* **2021**.

50. Li, R., Wang, S., Zhu, F. and Huang, J. In *Adaptive graph convolutional neural networks*, Proceedings of the AAAI Conference on Artificial Intelligence, 2018.

51. Wang, X., Zhu, M., Bo, D., Cui, P., Shi, C. and Pei, J. In *Am-gcn: Adaptive multi-channel graph convolutional networks*, Proceedings of the 26th ACM SIGKDD International conference on knowledge discovery & data mining, 2020; pp 1243-1253.

52. Chen, Y., Wu, L. and Zaki, M., Iterative deep graph learning for graph neural networks: Better and robust node embeddings. *Advances in Neural Information Processing Systems* **2020,** *33*, 19314-19326.

53. Deng;, A. and Hooi., B., Graph Neural Network-Based Anomaly Detection in Multivariate Time

Series. *aaai2021* **2021**.

54. Yu, D., Zhang, R., Jiang, Z., Wu, Y. and Yang, Y. In *Graph-revised convolutional network*, Joint European Conference on Machine Learning and Knowledge Discovery in Databases, Springer: 2020; pp 378-393.

55. Luo, D., Cheng, W., Yu, W., Zong, B., Ni, J., Chen, H. and Zhang, X. In *Learning to drop: Robust graph neural network via topological denoising*, Proceedings of the 14th ACM international conference on web search and data mining, 2021; pp 779-787.

56. Sun, Q., Li, J., Peng, H., Wu, J., Fu, X., Ji, C. and Yu, P. S., Graph Structure Learning with Variational Information Bottleneck. *arXiv preprint arXiv:2112.08903* **2021**.

57. Gao, X., Hu, W. and Guo, Z. In *Exploring structure-adaptive graph learning for robust semi-supervised classification*, 2020 IEEE International Conference on Multimedia and Expo (ICME), IEEE: 2020; pp 1-6.

58. Kaewprapha, P., Prempaneerach, P., Singh, V., Tinikul, T., Intarangsi, N. and Kijkanjanarat, T. In *Predicting Full Load, Partial Load Efficiency of a Combined Cycle Power Plant Using Machine Learning Methods*, 2022 7th International Conference on Computer and Communication Systems (ICCCS), IEEE: 2022; pp 11-16.

59. Vaswani, A., Shazeer, N., Parmar, N., Uszkoreit, J., Jones, L., Gomez, A. N., Kaiser, L. and Polosukhin, I., Attention Is All You Need. *arXiv* **2017**.

60. Ren, H., Xu, B., Wang, Y., Yi, C., Huang, C., Kou, X., Xing, T., Yang, M., Tong, J. and Zhang, Q. In *Time-series anomaly detection service at microsoft*, Proceedings of the 25th ACM SIGKDD international conference on knowledge discovery & data mining, 2019; pp 3009-3017.

61. Goh, J., Adepu, S., Junejo, K. N. and Mathur, A., A Dataset to Support Research in the Design of Secure Water Treatment Systems. *Springer, Cham* **2016**.

62. Ahmed, C. M., Palleti, V. R. and Mathur, A. P. In *WADI: a water distribution testbed for research in the design of secure cyber physical systems*, the 3rd International Workshop, 2017.

63. Zhan, J., Wang, S., Ma, X., Wu, C., Yang, C., Zeng, D. and Wang, S. In *Stgat-Mad: Spatial-Temporal Graph Attention Network For Multivariate Time Series Anomaly Detection*, ICASSP 2022-2022 IEEE International Conference on Acoustics, Speech and Signal Processing (ICASSP), IEEE: 2022; pp 3568-3572.

64. Park, D., Hoshi, Y. and Kemp, C. C., A Multimodal Anomaly Detector for Robot-Assisted Feeding Using an LSTM-based Variational Autoencoder. *IEEE Robotics and Automation Letters* **2017**, *PP* (99).

65. Audibert, J., Michiardi, P., Guyard, F., Marti, S. and Zuluaga, M. A. In *Usad: Unsupervised anomaly detection on multivariate time series*, Proceedings of the 26th ACM SIGKDD International Conference on Knowledge Discovery & Data Mining, 2020; pp 3395-3404.

66. Camacho, J., Pérez-Villegas, A., García-Teodoro, P. and Maciá-Fernández, G., PCA-based multivariate statistical network monitoring for anomaly detection. *Computers & Security* **2016**, *59*, 118-137.

67. Fei, T. L., Kai, M. T. and Zhou, Z. H. In *Isolation Forest*, IEEE International Conference on Data Mining, 2008.

68. Qi, M. In *LightGBM: A Highly Efficient Gradient Boosting Decision Tree*, Neural Information Processing Systems, 2017.

69. Hundman, K., Constantinou, V., Laporte, C., Colwell, I. and Soderstrom, T. In *Detecting Spacecraft Anomalies Using LSTMs and Nonparametric Dynamic Thresholding*, the 24th ACM SIGKDD International Conference, 2018.

70. Shen, L., Li, Z. and Kwok, J., Timeseries anomaly detection using temporal hierarchical one-class network. *Advances in Neural Information Processing Systems* **2020,** *33*, 13016-13026.

71. Kim, S., Choi, K., Choi, H. S., Lee, B. and Yoon, S., Towards a Rigorous Evaluation of Time-series Anomaly Detection. **2021**.